\documentclass[runningheads]{llncs}
\usepackage[T1]{fontenc}
\usepackage{graphicx}
\usepackage{amsmath,amssymb,amsfonts}
\begin{document}
\title{DyTTP: Trajectory Prediction with Normalization-Free Transformers}
%
%


\author{JianLin Zhu\inst{1} \and
HongKuo Niu\inst{2}
}
\authorrunning{Z. JianLin, N. HongKuo}
%
\institute{Shanghai Institute of Technology, Shanghai 10259, China\\
\email{877535678@qq.com} \\
\email{236142132@mail.sit.edu.cn}\\
}

\maketitle              
\begin{abstract}
Accurate trajectory prediction is a cornerstone for the safe operation of autonomous driving systems, where understanding the dynamic behavior of surrounding agents is crucial. Transformer-based architectures have demonstrated significant promise in capturing complex spatio-temporality dependencies. However, their reliance on normalization layers can lead to computation overhead and training instabilities. In this work, we present a two-fold approach to address these challenges. First, we integrate DynamicTanh (DyT), which is the latest method to promote transformers, into the backbone, replacing traditional layer normalization. This modification simplifies the network architecture and improves the stability of the inference. We are the first work to deploy the DyT to the trajectory prediction task. Complementing this, we employ a snapshot ensemble strategy to further boost trajectory prediction performance. Using cyclical learning rate scheduling, multiple model snapshots are captured during a single training run. These snapshots are then aggregated via simple averaging at inference time, allowing the model to benefit from diverse hypotheses without incurring substantial additional computational cost. Extensive experiments on Argoverse datasets demonstrate that our combined approach significantly improves prediction accuracy, inference speed and robustness in diverse driving scenarios. This work underscores the potential of normalization-free transformer designs augmented with lightweight ensemble techniques in advancing trajectory forecasting for autonomous vehicles.

\end{abstract}
\section{Introduction}
Accurate trajectory prediction is a cornerstone for the safe operation of autonomous driving systems, 
where the ability to forecast the future positions of surrounding agents is critical for timely decision-making and collision avoidance. 
Subsequent methods shifted toward vector-based representations to more accurately encode road geometry and agent interactions\cite{3}\cite{4}. Traditional sequence models such as LSTMs\cite{lstm} were initially popular for capturing temporal dependencies, however, their inherent limitations in parallelization and long-term dependency modeling have motivated the exploration of alternative architectures. 
Recent advances have seen the emergence of transformer-based methods that leverage self-attention\cite{self attention}(cite:self attention) to capture complex spatial-temporal dependencies across multiple agents. 
Models like HiVT\cite{hivt} have demonstrated impressive performance in trajectory prediction by modeling intricate relationships between agents and their environments. 
Despite these successes, the standard transformer architecture depends heavily on normalization 
layers—particularly Layer Normalization—which, while effective in stabilizing training, can impose additional computational overhead and sometimes lead to training instabilities in real-time applications\cite{layernorm}.\par

Despite these advantages, conventional transformer models typically rely on normalization layers—such as Layer Normalization—to stabilize training and improve convergence. However, such normalization techniques can introduce 
redundant computational and inference time and may lead to training instabilities when deployed in real-time systems. Recently, a novel DynamicTanh layer is proposed in \cite{dyt}, aiming to create a brand new transformer framework and achieve excellent performance on computer vision, large language models, diffusion models and many other tasks. Despite its potential in these work, DyT remains underexplored in the context of transformer-based trajectory prediction.\par

Building on these ideas, we propose a two-part approach to simplify the transformer backbone and 
improve inference speed while maintaining prediction accuracy.
First, inspired by DyT\cite{dyt}, we integrate DynamicTanh —a recent advancement in normalization-free transformer architectures—into the backbone, replacing traditional Layer Normalization. 
This modification streamlines the network structure and improves training stability by enabling more robust gradient propagation. 
Second, we adopt a snapshot ensemble strategy\cite{snap} that leverages cyclical learning rate scheduling to capture diverse model snapshots during a single training run, which is different from the tradictional ensembling technologies such as Bootstrap Aggregating\cite{boot} and Boosting\cite{boosting}. 
These snapshots are aggregated via simple averaging during every few inference cycles but not the whole process, allowing the model to benefit from multiple hypotheses while keeping the deployment overhead minimal.\par

Our approach is evaluated on the Argoverse dataset\cite{argoverse}, where extensive experiments demonstrate significant improvements in prediction accuracy, inference speed, and robustness across varied driving scenarios. By combining normalization-free transformer designs with a lightweight ensemble technique, our method offers a scalable and reliable solution for real-time trajectory forecasting, a critical requirement for advancing autonomous driving systems.\par

In summary, our contributions are as follows:
\renewcommand{\labelitemi}{\textbullet}
\begin{itemize}
\item We propose a normalization-free transformer backbone by partially replacing conventional Layer Normalization with DynamicTanh (DyT), resulting in a simpler and more stable network. We are the first work to disscus the application of DyT in the trajectory prediction task.
\item We introduce a snapshot ensemble strategy that captures multiple model snapshots during training and aggregates them at inference, enhancing prediction performance with minimal computational overhead.
\item Our extensive evaluation on the Argoverse dataset confirms that the proposed approach significantly improves trajectory prediction accuracy, inference speed, and robustness in diverse driving conditions.
\end{itemize}
\begin{figure}[t]
  \centering
  \includegraphics[width=\textwidth]{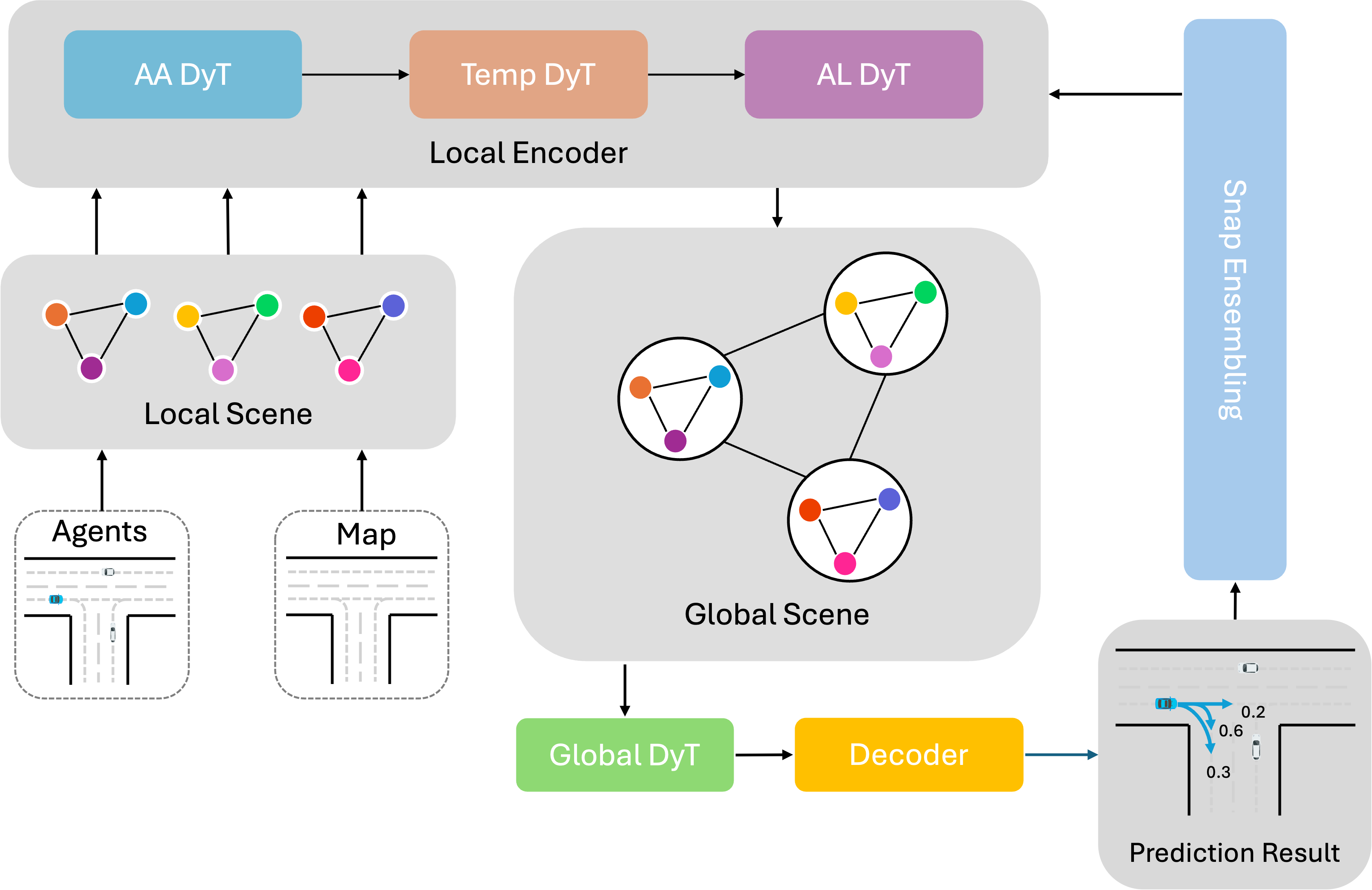}
  \caption{Overview of our DyTTP. AA DyT, Temp DyT, AL DyT and Global DyT denote agent-agnet, temporal, agent-lane and global interaction with normalization-free transformers, respectively.}
  \label{fig1}
\end{figure}

\section{Related Work}
The development of accurate and efficient trajectory prediction models is critical for autonomous driving, as they allow anticipating the future states of traffic agents to ensure safety and operational stability for real-time decisions.
\subsection{Trajectory Prediction with Transformers}
Transformer-based architectures have rapidly emerged as a leading approach for modeling the social spatial and temporal interactions between agents and agents, agents and lanes because of their self-attention mechanisms, which enables the modeling of long-range dependencies and complex interactions. Early approaches based on recurrent neural networks (RNNs)\cite{rnn,1,33} and LSTMs\cite{lstm1,lstm2,lstm3,lstm4} faced challenges in parallelization and capturing long-term dependencies, prompting researchers to explore transformer alternatives. For instance, methods such as HiVT\cite{hivt}, HPNet\cite{hpnet} and QCNet\cite{qcnet} have demonstrated that leveraging transformer architectures can significantly improve the modeling of spatio-temporal relationships and multi-modal trajectory outputs. Despite these successes, the reliance on traditional normalization layers—most notably Layer Normalization—remains a challenge. These layers, while stabilizing training, add extra computational overhead and may sometimes lead to training instabilities, especially in real-time prediction tasks. Recent innovations, such as the integration of DyT, seek to overcome these limitations by promoting normalization-free transformer architectures that maintain high performance while reducing computational demands.
\subsection{Trajectory Prediction with Ensembling Technology}
Ensemble techniques have been widely adopted in machine learning as a means to boost model performance and enhance robustness by combining the predictions of multiple models. In trajectory prediction, the inherent uncertainty and multi-modality of future agent behaviors make ensembling an attractive strategy for generating diverse hypotheses. Traditional ensembling methods—such as bagging or boosting—typically require training several separate models, which can be computationally expensive\cite{hpnet1,hpnet2}. MultiPath++\cite{multipath++} uses ensembling of multiple learned heads or multiple trajectory samples to improve diversity and reduce prediction errors, which can lead to redundant computation. We choose an effective alternative, called snapshot ensembling, where a single training run is used to capture multiple model snapshots via a cyclical learning rate schedule. Each snapshot, representing a different local optimum, contributes to a robust aggregate prediction through simple averaging at inference time. Snapshot ensembling has proven effective in fields like computer vision and natural language processing\cite{snapshot,google}, but its application to trajectory prediction remains relatively underexplored. By integrating snapshot ensembling into a transformer-based trajectory prediction framework, our work harnesses the benefits of diverse model hypotheses while keeping the deployment complexity minimal.
\section{Method}
In this section, we describe our proposed approach for improving trajectory prediction using normalization-free transformers and snapshot ensembling. The whole framework shown in ~\ref{fig1}. We first introduce the backbone architecture, which integrates DyT to replace traditional Layer Normalization. We then detail the snapshot ensemble strategy, which captures multiple model snapshots during training and aggregates them at inference time. Finally, we present the training loss function used to optimize the model parameters.
\subsubsection{Problem Formulation}
Given a sequence of observed agent trajectories and lane positions $\mathbf{X} = \{x_1, x_2, ..., x_T\}$, where $x_t = (x_t^1, x_t^2, ..., x_t^N, l_t^1, l_t^2, ..., l_t^M)$ represents the $N$ agents' positions and $M$ lane positions at time step $t$, the goal of trajectory prediction is to forecast the future positions $\hat{\mathbf{Y}} = \{\hat{y}_{T+1}, \hat{y}_{T+2}, ..., \hat{y}_{T+F}\}$ of these agents over a prediction horizon $F$. Each predicted trajectory $\hat{y}_t = (\hat{y}_t^1, \hat{y}_t^2, ..., \hat{y}_t^N)$ consists of the $N$ agents' positions at time step $t$.
\begin{figure}[t]
  \centering
  \includegraphics[width=\textwidth,height=0.4\textwidth]{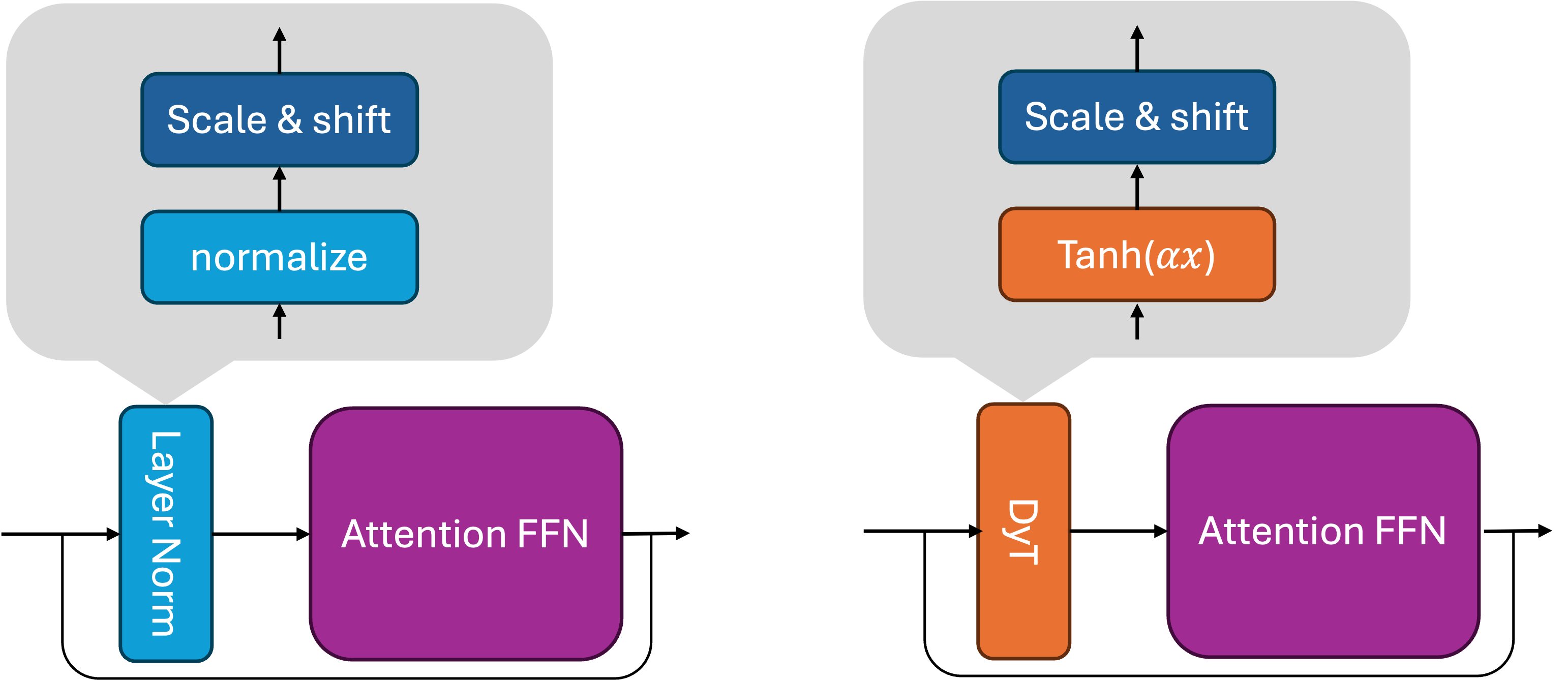}
  \caption{Original transformer block(left) and DynamicTanh layer(right), which is a straighforward repalcement for traditional Layer Normalization.}
  \label{fig2}
\end{figure}

\subsubsection{Backbone}
We employ a typical transformer-based model, HiVT, as our backbone architecture. This model consists of an encoder-decoder structure with self-attention mechanisms to capture spatial and temporal dependencies between agents and lanes. To enhance the stability and efficiency of the model, we replace the traditional Layer Normalization layers in the encoder and decoder with DyT layers. DyT is a normalization-free activation function that adaptively scales the input features based on the mean and variance of the input tensor. This modification simplifies the network architecture and improves the stability of the training process by enabling more robust gradient propagation. The backbone architecture is illustrated in ~\ref{fig1}.
\subsubsection{DynamicTanh}
The DynamicTanh (DyT) layer is a normalization-free activation function that adaptively scales the input features as shown in ~\ref{fig2}. This scaling operation is performed using a dynamic scaling factor $\alpha$ and a dynamic shift factor $\beta$, which are learned during training. The DyT layer is defined as follows:
\begin{equation}
\text{DyT}(\textbf{x}) = \gamma \cdot \text{tanh}(\alpha \textbf{x}) + \beta
\end{equation}
where $x$ is the input tensor, $\alpha$ is a learnable scalar paramete, $\beta$ and $\gamma$ are learnable, per-channel vector parameters, respectively. By replacing traditional Layer Normalization with DyT, we improve the stability of the training process and the result of experiments shows that the inference speed is also improved, which we will explain in section4(need cite).
\subsubsection{Snapshot Ensemble}
To further enhance the prediction performance, we adopt a snapshot ensemble strategy that captures multiple model snapshots during training and aggregates them at inference time. This strategy leverages cyclical learning rate scheduling to capture diverse model hypotheses, which are then combined to produce a robust aggregate prediction. Specifically, we use a cosine annealing learning rate schedule to train the model, capturing a snapshot of the model parameters at the end of each learning rate cycle. It is defined as follows:
\begin{equation}
\eta_i = \eta_{min} + \frac{1}{2}(\eta_{max} - \eta_{min})(1 + \cos(\frac{E_{cur}}{E_i}\pi))
\end{equation}
where $\eta_i$ is the learning rate at iteration $i$, $\eta_{min}$ and $\eta_{max}$ are the minimum and maximum learning rates, $E_{cur}$ is the current epoch since the last restart, and $E_i$ is the number of epochs between two warm restarts. By capturing multiple model snapshots during training, we can generate diverse hypotheses that improve the prediction accuracy without incurring substantial additional computational cost.\par
These snapshots are then aggregated via simple averaging at inference time, allowing the model to benefit from diverse hypotheses without incurring substantial additional computational cost.
\subsubsection{Training Loss}
The training loss function consists of two components: a regression loss for trajectory prediction and a classification loss for multi-modal prediction. The total loss is defined as:
\begin{equation}
\mathcal{L} = \mathcal{L}_{\text{reg}} + \lambda \mathcal{L}_{\text{cls}}
\end{equation}
where $\mathcal{L}_{\text{reg}}$ is the regression loss, $\mathcal{L}_{\text{cls}}$ is the classification loss, and $\lambda$ is a weighting factor to balance the two losses.

The regression loss $\mathcal{L}_{\text{reg}}$ is based on the negative log-likelihood of the predicted trajectories:
\begin{equation}
\mathcal{L}_{\text{reg}} = -\sum_{i=1}^{N} \log p(\mathbf{y}_i | \mathbf{x}_i, \theta)
\end{equation}
where $p(\mathbf{y}_i | \mathbf{x}_i, \theta)$ represents the likelihood of the ground truth trajectory $\mathbf{y}_i$ given the input $\mathbf{x}_i$ and model parameters $\theta$.

The classification loss $\mathcal{L}_{\text{cls}}$ uses cross-entropy to optimize the probabilities of selecting the correct trajectory mode:
\begin{equation}
\mathcal{L}_{\text{cls}} = -\sum_{i=1}^{N} \sum_{k=1}^{K} y_{i,k} \log \hat{y}_{i,k}
\end{equation}
where $y_{i,k}$ is the ground truth one-hot vector for the $k$-th mode, and $\hat{y}_{i,k}$ is the predicted probability for the $k$-th mode.

By combining these two losses, the model is trained to accurately predict both the trajectories and their associated probabilities.

\begin{table}[t]
  \centering
  \begin{tabular}{l|c c c }
    \hline
    Method & minADE & minFDE & MR \\
    \hline
    LaneGCN\cite{4} & 0.8679 & 1.3640 & 0.1634 \\
    Scene Transformer\cite{scene} & 0.8026 & 1.2321 & 0.1255 \\
    DenseTNT\cite{DenseTNT} & 0.8817 & 1.2815 & 0.1258 \\
    MutltiModalTransformer\cite{multi} & 0.8372 & 1.2905 & 0.1429 \\
    mmTransformer\cite{mm} & 0.8436 & 1.3383 & 0.1540 \\
    HOME+GOME\cite{home,gome} & 0.8904 & 1.2919 & 0.0846 \\
    TPCN\cite{tpcn} & 0.8153 & 1.2442 & 0.1333 \\
    \hline
    \textbf{DyTTP(ours)} & \textbf{0.7845} & \textbf{1.1948} & \textbf{0.1331} \\
    \hline
  \end{tabular}
  \caption{Comparison of prediction performance on the Argoverse test set.}
  \label{table1}
\end{table}

\section{Experiments}
In this section, we present the experimental setup and results to evaluate the effectiveness of our proposed approach. We first describe the dataset and evaluation metrics used in our experiments. We then compare our method with several baselines to demonstrate its superiority. Finally, we conduct an ablation study to analyze the impact of each component of our approach on the prediction performance.
\subsubsection{Dataset}
To evaluate the performance of our model, we conduct experiments on the Argoverse dataset\cite{argoverse}. This dataset contains high-definition maps and sensor data collected from autonomous vehicles in various urban environments. The dataset includes annotated trajectories of vehicles, pedestrians, and cyclists, as well as lane information and traffic signals. We use the Argoverse forecasting benchmark, which consists of 205,942 training samples and 39,472 validation samples. Each sample contains the observed trajectories of agents and lanes, as well as the future trajectories to be predicted. The goal is to predict the future positions of agents over a prediction horizon of 30 time steps based on 20 time steps historical data.
\subsubsection{Metrics}
We evaluate the prediction performance using two standard metrics: the minimum Average Displacement Error (minADE), the minimum Final Displacement Error (minFDE) and the Miss Rate(MR). ADE measures the average Euclidean distance between the predicted and ground truth trajectories over the prediction horizon, while FDE measures the Euclidean distance at the final time step. MR calculates the proportion of predicted trajectories whose endpoints deviate more than 2.0 merters from the actual ground truth endpoint. Lower values of ADE, FDE and MR indicate better prediction accuracy.\par
Additionally, we also evaluate the inference speed and robustness of our model, which is measured in millisecond(ms), on the Argoverse Validation set. A shorter time indicates faster prediction speed and better real-time performance, a shorter standard deviation indicates better robustness.

\begin{table}[t]
  \centering
  \begin{tabular}{c|c|c}
    \hline
    Metrics & HiVT\cite{hivt} & \textbf{DyTTP(ours)} \\
    \hline
    ave(ms) & \textbf{42.56} & 43.47 \\
    std(ms) & 9.93 & \textbf{9.82} \\
    min(ms) & 27.5 & \textbf{26.62} \\
    max(ms) & \textbf{269.76} & 272.24 \\
    ADE & 0.66 & 0.66 \\
    FDE & \textbf{0.96} & 0.98 \\
    MR & 0.09 & 0.09 \\
    \hline
  \end{tabular}
  \caption{Comparison of inference speed and prediction performance.}
  \label{table2}
\end{table}

\begin{table}[t]
  \centering
  \begin{tabular}{c c c|c c c c}
    \hline
    DyT & Snapshot & Backbone & ADE & FDE & MR & inf(ms)\\
    \hline
     & & \checkmark & \textbf{0.66} & \textbf{0.96} & 0.09 & 27.5\\
    \checkmark & & \checkmark & 0.67 & 0.99 & 0.09 & 27.16\\
      & \checkmark & \checkmark & 0.67 & 0.98 & 0.09 & 27.27\\
    \checkmark & \checkmark & \checkmark & \textbf{0.66} & 0.98 & 0.09 & \textbf{26.62} \\
    \hline
  \end{tabular}
  \caption{Ablation results on Arogverse validation set. Inf denotes the inference speed.}
  \label{table3}
\end{table}

\subsubsection{Comparison with Baselines}
The results of marginal trajectory prediction on the Argoverse dataset are presented in ~\ref{table1}. Our DyTTP achieves competitive results in terms of minADE, minFDE and MR compared to the baseline models. Results in ~\ref{table2} show that our DyTTP achieves faster maximum inference speeds and better inference stability while maintaining almost the same performance.
\subsubsection{Ablation Study}
To analyze the impact of each component of our approach on the prediction performance, we conduct an ablation study on the Argoverse dataset. We compare the performance of our full model with the variants that exclude DyT and Snapshot Ensembling. The results in ~\ref{table3} demonstrate that both DyT and Snapshot Ensembling contribute to the improvement in prediction accuracy. The combination of these two components achieves the best performance across all metrics.
\subsubsection{Effect of DynamicTanh}
Results in ~\ref{table3} show that the DyT layer improves the inference speed and maintians the MR at 0.09, which is the same as the baseline model. The ADE and FDE are slightly worse than the baseline model, but the difference is negligible. This indicates that the DyT layer can effectively replace traditional Layer Normalization and improve the stability of the training process without sacrificing prediction accuracy.
\subsubsection{Effect of Snapshot Ensembling}
The results in ~\ref{table3} also show that Snapshot Ensembling improves the inference speed without excessive loss in ADE and FDE. The MR remains the same as the baseline model, indicating that Snapshot Ensembling can effectively capture diverse model hypotheses and improve the prediction performance without incurring substantial additional computational cost.
\section{Conclusion}
In this paper, we introduce the latest normalization-free transformer architecture, DyT, into the trajectory prediction task. We propose a two-fold approach to improve the prediction accuracy, inference speed, and robustness of trajectory forecasting models. First, we integrate DyT into the transformer backbone to replace traditional Layer Normalization, simplifying the network architecture and improving training stability. Second, we adopt a snapshot ensemble strategy to capture diverse model snapshots during training and aggregate them at inference time, enhancing prediction performance with minimal computational overhead. Extensive experiments on the Argoverse dataset demonstrate that our combined approach significantly improves prediction accuracy, inference speed, and robustness in diverse driving scenarios. Our work underscores the potential of normalization-free transformer designs augmented with lightweight ensemble techniques in advancing trajectory forecasting for autonomous vehicles.

%
%

%
%
%

\begin{thebibliography}{8}
\bibitem{3}
J. Gao, C. Sun, H. Zhao, Y. Shen, D. Anguelov, C. Li, and C. Schmid,
“Vectornet: Encoding hd maps and agent dynamics from vectorized rep-
resentation,” in Proceedings of the IEEE/CVF conference on computer
vision and pattern recognition, pp. 11525–11533, 2020.
\bibitem{4}
M. Liang, B. Yang, R. Hu, Y. Chen, R. Liao, S. Feng, and R. Urtasun,
“Learning lane graph representations for motion forecasting,” in Com-
puter Vision–ECCV 2020: 16th European Conference, Glasgow, UK,
August 23–28, 2020, Proceedings, Part II 16, pp. 541–556, Springer,
2020.
\bibitem{lstm}
Graves, Alex, and Alex Graves. "Long short-term memory." Supervised sequence labelling with recurrent neural networks (2012): 37-45.
\bibitem{self attention}
Vaswani, Ashish, et al. "Attention is all you need." Advances in neural information processing systems 30 (2017).
\bibitem{hivt}
Z. Zhou, L. Ye, J. Wang, K. Wu, and K. Lu, “Hivt: Hierarchical vector transformer for multi-agent motion prediction,” in Proceedings of the IEEE/CVF Conference on Computer Vision and Pattern Recognition, pp. 8823–8833, 2022.
\bibitem{layernorm}
Ba, Jimmy Lei, Jamie Ryan Kiros, and Geoffrey E. Hinton. "Layer normalization." arxiv preprint arxiv:1607.06450 (2016).
\bibitem{dyt}
Zhu, Jiachen, et al. "Transformers without Normalization." arxiv preprint arxiv:2503.10622 (2025).
\bibitem{snap}
Shaban, Ali, and Heiko Paulheim. "SnapE–training snapshot ensembles of link prediction models." International Semantic Web Conference. Cham: Springer Nature Switzerland, 2024.
\bibitem{boot}
Lee, Tae-Hwy, Aman Ullah, and Ran Wang. "Bootstrap aggregating and random forest." Macroeconomic forecasting in the era of big data: Theory and practice (2020): 389-429.
\bibitem{boosting}
Margineantu, Dragos D., and Thomas G. Dietterich. "Pruning adaptive boosting." ICML. Vol. 97. 1997.
\bibitem{argoverse}
M.-F. Chang, J. Lambert, P. Sangkloy, J. Singh, S. Bak, A. Hartnett,

D. Wang, P. Carr, S. Lucey, D. Ramanan, et al., “Argoverse: 3d tracking and forecasting with rich maps,” in Proceedings of the IEEE/CVF conference on computer vision and pattern recognition, pp. 8748–8757, 2019.
\bibitem{rnn}
R. M. Schmidt, “Recurrent neural networks (rnns): A gentle introduction and overview,” arXiv preprint arXiv:1912.05911, 2019.
\bibitem{lstm1}
X. Chen, H. Zhang, F. Zhao, Y. Hu, C. Tan, and J. Yang, “Intentionaware vehicle trajectory prediction based on spatial-temporal dynamic attention network for internet of vehicles,” IEEE Transactions on Intelligent Transportation Systems, vol. 23, no. 10, pp. 19471–19483, 2022.
\bibitem{lstm2}
Y. Xing, C. Lv, and D. Cao, “Personalized vehicle trajectory prediction based on joint time-series modeling for connected vehicles,” IEEE Transactions on Vehicular Technology, vol. 69, no. 2, pp. 1341–1352, 2019.
\bibitem{lstm3}
A. Alahi, K. Goel, V. Ramanathan, A. Robicquet, L. Fei-Fei, and

S. Savarese, “Social lstm: Human trajectory prediction in crowded spaces,” in Proceedings of the IEEE conference on computer vision and pattern recognition, pp. 961–971, 2016.
\bibitem{lstm4}
N. Deo and M. M. Trivedi, “Convolutional social pooling for vehicle trajectory prediction,” in Proceedings of the IEEE conference on computer vision and pattern recognition workshops, pp. 1468–1476, 2018.
\bibitem{qcnet}
Z. Zhou, J. Wang, Y.-H. Li, and Y.-K. Huang, “Query-centric trajectory prediction,” in Proceedings of the IEEE/CVF Conference on Computer Vision and Pattern Recognition, pp. 17863–17873, 2023.
\bibitem{hpnet}
X. Tang, M. Kan, S. Shan, Z. Ji, J. Bai, and X. Chen, “Hpnet: Dynamic trajectory forecasting with historical prediction attention,” in Proceedings of the IEEE/CVF Conference on Computer Vision and Pattern Recognition, pp. 15261–15270, 2024.
\bibitem{hpnet1}
Xishun Wang, Tong Su, Fang Da, and Xiaodong Yang. Prophnet: Efficient agent-centric motion forecasting with anchor-informed proposals. In Proceedings of the IEEE/CVF Conference on Computer Vision and Pattern Recognition (CVPR), 2023. 1, 2, 3, 5, 6
\bibitem{hpnet2}
Maosheng Ye, Jiamiao Xu, Xunnong Xu, Tengfei Wang, Tongyi Cao, and Qifeng Chen. Dcms: Motion forecasting with dual consistency and multi-pseudo-target supervision. arXiv preprint arXiv:2204.05859, 2022. 2, 6, 8
\bibitem{multipath++}
B. Varadarajan, A. Hefny, A. Srivastava, K. S. Refaat, N. Nayakanti,

A. Cornman, K. Chen, B. Douillard, C. P. Lam, D. Anguelov, et al., “Multipath++: Efficient information fusion and trajectory aggregation for behavior prediction,” in 2022 International Conference on Robotics and Automation (ICRA), pp. 7814–7821, IEEE, 2022.
\bibitem{snapshot}
Huang, Gao, et al. "Snapshot ensembles: Train 1, get m for free." arxiv preprint arxiv:1704.00109 (2017).
\bibitem{google}
Wu, Yonghui, et al. "Google's neural machine translation system: Bridging the gap between human and machine translation." arxiv preprint arxiv:1609.08144 (2016).
\bibitem{1}
N. Lee, W. Choi, P. Vernaza, C. B. Choy, P. H. Torr, and M. Chandraker, “Desire: Distant future prediction in dynamic scenes with interacting agents,” in Proceedings of the IEEE conference on computer vision and pattern recognition, pp. 336–345, 2017.
\bibitem{33}
A. Zyner, S. Worrall, and E. Nebot, “Naturalistic driver intention and path prediction using recurrent neural networks,” IEEE transactions on intelligent transportation systems, vol. 21, no. 4, pp. 1584–1594, 2019.
\bibitem{scene}
Jiquan Ngiam, Benjamin Caine, Vijay Vasudevan, Zhengdong Zhang, Hao-Tien Lewis Chiang, Jeffrey Ling, Rebecca Roelofs, Alex Bewley, Chenxi Liu, Ashish Venugopal, et al. Scene transformer: A uniﬁed architecture for predicting multiple agent trajectories. In Proceedings of the International Conference on Learning Representations (ICLR), 2022. 1, 2, 7
\bibitem{DenseTNT}
Junru Gu, Chen Sun, and Hang Zhao. Densetnt: End-to-end trajectory prediction from dense goal sets. In Proceedings of the IEEE/CVF International Conference on Computer Vision (ICCV), 2021. 2, 7, 8
\bibitem{multi}
Zhiyu Huang, Xiaoyu Mo, and Chen Lv. Multi-modal motion prediction with transformer-based neural network for autonomous driving. arXiv preprint arXiv:2109.06446, 2021.
\bibitem{mm}
Yicheng Liu, Jinghuai Zhang, Liangji Fang, Qinhong Jiang, and Bolei Zhou. Multimodal motion prediction with stacked transformers. In Proceedings of the IEEE/CVF Conference on Computer Vision and Pattern Recognition (CVPR), 2021.
\bibitem{home}
Thomas Gilles, Stefano Sabatini, Dzmitry Tsishkou, Bogdan Stanciulescu, and Fabien Moutarde. Gohome: Graphoriented heatmap output forfuture motion estimation. arXiv preprint arXiv:2109.01827, 2021.
\bibitem{gome}
Thomas Gilles, Stefano Sabatini, Dzmitry Tsishkou, Bogdan Stanciulescu, and Fabien Moutarde. Home: Heatmap output for future motion estimation. In IEEE International Conference on Intelligent Transportation Systems (ITSC), 2021.
\bibitem{tpcn}
Maosheng Ye, Tongyi Cao, and Qifeng Chen. Tpcn: Temporal point cloud networks for motion forecasting. In Proceedings of the IEEE/CVF Conference on Computer Vision and Pattern Recognition (CVPR), 2021.

\end{thebibliography}
%

\end{document}